\documentclass[Afour,sagev,times]{sagej}

\usepackage{moreverb,url}

\usepackage[colorlinks,bookmarksopen,bookmarksnumbered,citecolor=red,urlcolor=red]{hyperref}

\newcommand\BibTeX{{\rmfamily B\kern-.05em \textsc{i\kern-.025em b}\kern-.08em
T\kern-.1667em\lower.7ex\hbox{E}\kern-.125emX}}

\usepackage{epstopdf}
\usepackage{graphicx}
\usepackage{subfigure}
\usepackage{algpseudocode}
\usepackage{algorithm}
\usepackage{algorithmicx}
\usepackage{url}

\usepackage{xcolor}

\begin{document}

\runninghead{Hilasaca and Paulovich}

\title{Visual Feature Fusion and its Application to Support Unsupervised Clustering Tasks}

\author{Gladys Hilasaca \affilnum{1} and Fernando Paulovich \affilnum{2}}

\affiliation{\affilnum{1}University of Sao Paulo, Brazil\\
\affilnum{2}Dalhousie University, Canada}

\corrauth{Fernando Paulovich,  Faculty of Computer Science,
Dalhousie University,	
6050 University Avenue,
Halifax, NS,
Canada.}

\email{paulovich@dal.ca}

\begin{abstract}

On visual analytics applications, the concept of putting the user on the loop refers to the ability to replace heuristics by user knowledge on machine learning and data mining tasks.  On supervised tasks, the user engagement occurs via the manipulation of the training data. However, on unsupervised tasks, the user involvement is limited to changes in the algorithm parametrization or the input data representation, also known as features. Depending on the application domain, different types of features can be extracted from the raw data. Therefore, the result of unsupervised algorithms heavily depends on the type of employed feature. Since there is no perfect feature extractor, combining different features have been explored in a process called feature fusion. The feature fusion is straightforward when the machine learning or data mining task has a cost function. However, when such a function does not exist, user support for combination needs to be provided otherwise the process is impractical. In this paper, we present a novel feature fusion approach that uses small data samples to allows users not only to effortless control the combination of different feature sets but also to interpret the attained results. The effectiveness of our approach is confirmed by a comprehensive set of qualitative and quantitative tests, opening up different possibilities of user-guided analytical scenarios not covered yet. The ability of our approach to providing real-time feedback for the feature fusion is exploited on the context of unsupervised clustering techniques, where the composed groups reflect the semantics of the feature combination. 
\end{abstract}


\maketitle
\section{Introduction}
\label{sec:intro}

Machine learning and data mining techniques are, in general, split into supervised and unsupervised approaches or a combination of both. On supervised approaches, user knowledge is added to the analytical process through sets of already analyzed instances. On unsupervised, knowledge can be added by changing algorithms' parameters or the input data representation, also known as features. The challenge is, therefore, not only to define the most appropriate set of parameters but also to find the data representation that best expresses the user or analyst knowledge.

Depending on the application domain (e.g., text or image), there exist several approaches to construct features, each providing complementary information about the original or raw data. Since there is no perfect feature, the idea of joining different representations is straightforward. This process is called data or feature fusion~\cite{Bostrom:2007}, and can occur through the combination of features (vector) or merging distances calculated from the features.  When the machine learning or data mining task involves a cost function, for instance, classification accuracy, such a function can be used to guide the combination. However, for tasks, like clustering~\cite{Xu:2005, Tan:2005} or multidimensional projection~\cite{Nonato:2018,Sacha:2017}, where such a function does not exist, user support for combination needs to be provided. Otherwise, in practice, the data fusion is impossible or useless given the abundance of different combinations. 

In this paper, we present a novel feature fusion approach that allows users to control and understand the fusion of different feature sets.  Starting with a small sample, users employ a simple widget to define the weights for the combination and observe the outcome in real-time through a scatterplot-based visualization. Once the user finds the weighted combination that best matches his or her point of view of similarity, the same weights are used to combine the complete dataset. In this way, we not only allow users to effortless test different combinations but also enables the interpretation of the attained results.

In summary, the main contributions of this paper are:
\begin{itemize}
    \item A novel feature fusion technique that allows users to explore and understand different combinations of features in real-time;
    \item An approach to input user knowledge into unsupervised tasks much more interpretable than parameter tweaking;
     \item An interactive visualization-assisted tool to explore large image collections that allows real-time tuning of the similarity between images to match users expectations.
\end{itemize}


\section{Related Work}
\label{sec:related}

The process of integrating information from multiple sources to produce a unified enhanced data model is called data fusion~\cite{Bostrom:2007}. The reason is to combine different data representations into a single model aiming at incorporating properties of the various sources. Data fusion can occur in different ways, including combining features, that is, the vectorial data representation, or merging distances calculated from the various sources.  

The concept of merging features is called feature fusion. Feature fusion aims at generating a unified vectorial data representation based on different sets of features (vectorial representations)~\cite{Utthara:2010, Sudha:2017}. The most straightforward approach is the feature concatenation~\cite{Anne:2015, KuangCLY:2016}. In the concatenation, given the sets of features $F_1,F_2,\ldots,F_p$, the unified representation is given by $[F_1,F_2,\ldots,F_p]$. Despite its simplicity, the literature reports several examples. In~\cite{Wang:2009}, Local Binary Pattern (LBP)~\cite{Ahonen:2004} and Histograms of Oriented Gradients (HOG)~\cite{Dalalhog} features are concatenated to improve performance in pedestrian detection. In~\cite{Manshor:2012}, Scale Invariant Transform Features (SIFT)~\cite{lowe:1999} and boundary-based shape~\cite{Gonzalez:2004} features are concatenated to improve object recognition. In~\cite{Chu:2018}, high, low, and medium-layers features of a deep neural network are united to support object detection, and in~\cite{Chun:2008} color and texture features are progressively concatenated aiming at reducing model complexity in a content-based retrieval framework. Feature concatenation was also used in the text domain. In~\cite{Loni:2011b}, the authors extract seven types of lexical, syntactical and semantic features and combine subsets of them to improve text classification. 

Weights can be used in the concatenation process to control the influence of the different features. In this process, the unified representations is given by $[\alpha_1F_1, \alpha_2F_2,\ldots,\alpha_pF_p]$~\cite{Utthara:2010}, where $\alpha_1, \alpha_2, \ldots,\alpha_p$ are the weights. In~\cite{Loni:2011}, the weighted concatenation was used to improve text classification by combining lexical, syntactic, and semantic features. In~\cite{Ma:2016}, a neural network was used to learn the weights of a concatenation, combining different image features, such as color, shape, and texture, to improve classification accuracy. In~\cite{You:2017}, the authors use a saliency detection model to fuse color and texture features through a weighting strategy. They first transform the color and texture features in saliency features and then linearly combine the saliency features. Different from the previous weighted techniques, in this case, they linearly combine the features instead of concatenating them, that is, the unified representation is given by $[\alpha_1F_1+ \alpha_2F_2+\ldots+\alpha_pF_p]$. This is possible since the saliency representations have the same dimensionality.
            
In practice, the feature concatenation is not recommended since it may result in a huge feature vectors leading to the curse of dimensionality problem~\cite{Utthara:2010}. One solution is to apply a dimensionality reduction after the concatenation~\cite{Yu:2017}, or to perform a distance fusion. In the distance fusion, instead of combining the vectorial representations, the distances calculated from the representations are combined. If $\Delta(F_i)$ represents the distance matrix calculated from $F_i$, the resulting distance matrix is given by $\alpha_1\Delta(F_1)+ \alpha_2\Delta(F_2)+\ldots+\alpha_p\Delta(F_p)$. In~\cite{Degani:2013}, a simple normalized combination of distances computed from different types of features is used to cover song identification. The distance fusion can also be performed using weights. In~\cite{Huang:2010}, weights are used to combine distances calculated from color and texture features to improve the results of a content-based image retrieval system. In~\cite{Vadivel:2004}, distances calculated from color and texture features are also combined to support content-based image retrieval applications. Finally, in~\cite{Liu:2017} and in~\cite{Chu:2018} distances calculated from features extracted from different layers of a deep neural network are combined seeking to improve retrieval tasks.

Different from data fusion, model fusion combines computational models instead of data. Such combination can be performed in two different ways: by combining different models (parametrizations) processing a single feature set (data set), or by combining different models processing different feature sets~\cite{Kim:2016}. The former is called ensemble learning and has been extensively used for classification tasks. The idea is to combine the prediction of different models using some voting strategy to improve model diversity and classification accuracy~\cite{Mendes-Moreira:2012, Dietterich:2000}. Ensembles of classifiers typically outperform single classifiers~\cite{Schneider:2017} and have been used in different domains, including remote sensing, computer security, financial risk assessment, fraud detection, recommender systems, medical computer-aided diagnosis, and others~\cite{Woniak:2014, Kim:2016}. Similarly, the later also employs a (weighted) voting strategy to combine different models, but in this case, the models use as input different sets of features. Examples of applications include fruit classification~\cite{KuangCLY:2016} and sentiment analysis~\cite{Xia:2011}.


Common to all these data and model fusion approaches is that the combinations can only be appropriately performed when a loss function is available to guide the process, like in classification. If such a function does not exists, or there is a degree of subjectivity in the process, the combination without proper user support hampers its applicability in practice or real scenarios, and none of the mentioned approaches offer such support. In this paper we devise an approach to aid on the process of feature combination, allowing users to control the process to match individual expectations,  enabling applications where the user judgment is crucial.

\section{Proposed Methodology}
\label{sec:method}

Our approach for feature fusion employs a two phase strategy to support users on defining combinations that reflect a particular point-of-view regarding similarity relationships. On the first phase, samples $S_1, S_2, \ldots, S_p$ are extract from each different set of features $F_1, F_2, \ldots, F_p$ and merged so that 
each set $S_i$ presents the same objects but represented using the different types of feature. Each sample $S_i$ is then mapped to a vectorial representation $R_i \in \mathbb{R}^m$ preserving as much as possible the distance relationships between the instances. These vectorial representations are then combined to generate a single representation $\overline{R}=\alpha_1R_1+\alpha_2R2+\ldots+\alpha_pR_p$, which is visualized. 

\begin{figure}[]
\centering
\includegraphics[width=\linewidth]{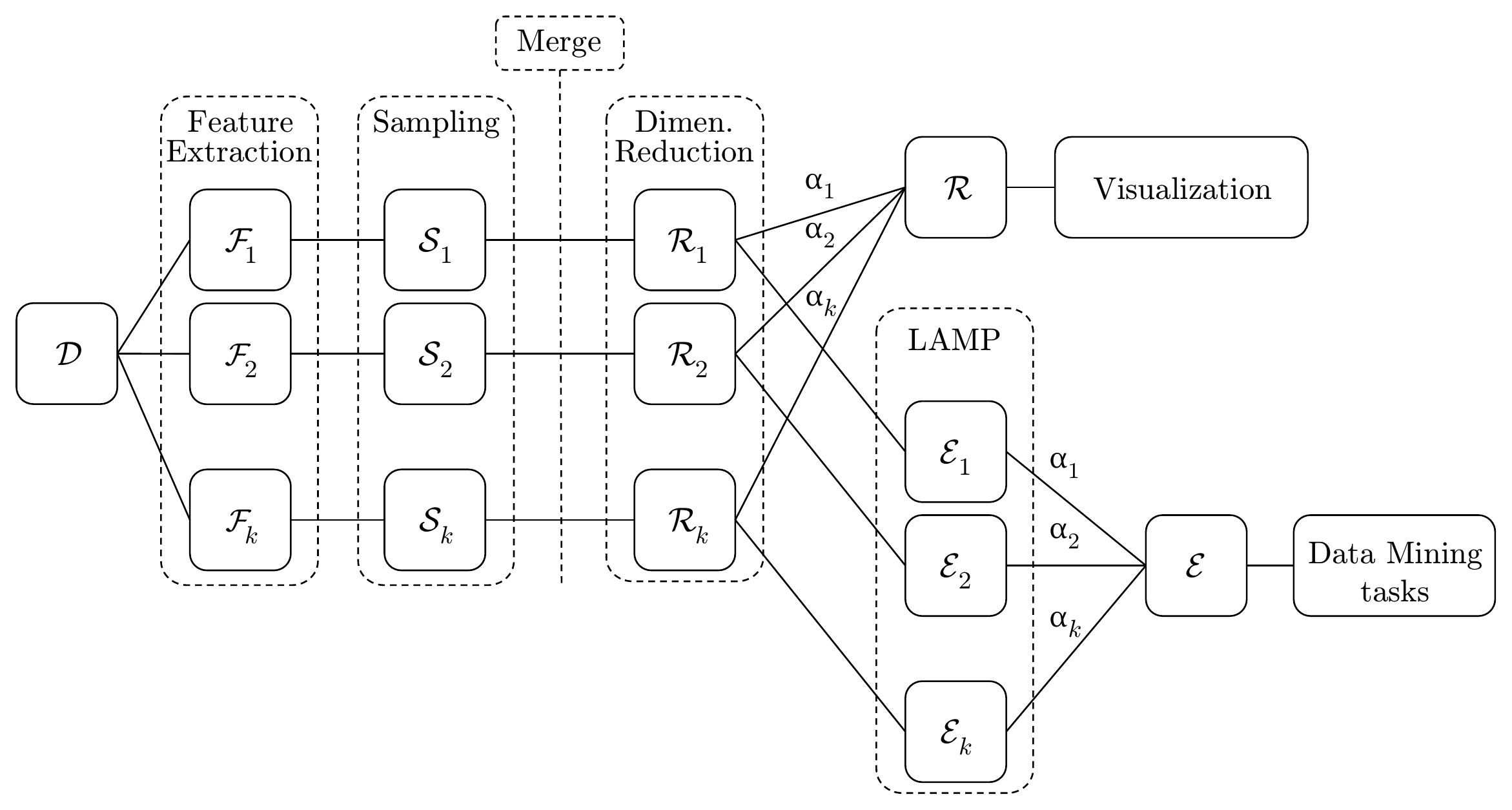}
\caption{Overview of our process for feature fusion. Initially a sample is extracted, combined and visualized. Based on that, the user can test different weights to fuse the features and observe the outcome. Once sample combination reflects the user expectation, the same weights are used to combine the complete sets of features that can them be used on subsequent tasks, such as clustering.}\label{fig:overview}\end{figure}

The user can then change the features weights and observe the outcome. Once the sample visualization reflects the user expectations, that is, once the proper weights $\alpha_1, \alpha_2, \ldots, \alpha_p$ are found, the second step takes place and the defined weights are used to combine the complete sets of features. In this process, the vectorial sample representations $R_1,R_2,\ldots,R_p$ and the samples $S_1, S_2, \ldots, S_p$ are used to construct models to map each set of feature $F_i$ to a vectorial representation $V_i \in \mathbb{R}^m$. Since these vectorial representations are embedded in the same space, they can be combined using the weights $\alpha_1, \alpha_2, \ldots, \alpha_p$, obtaining the final vectorial representation $\overline{V}$ that matches the users expectations defined by the sample visualization. Figure~\ref{fig:overview} outlines our approach showing the involved steps. Next we detail these steps, starting with the sampling and the dimensionality reduction.

\subsection{Sampling and Mapping}

The first step of our process is sampling. Since users employ the sample visualization to guide the feature fusion process, it is important to have all possible data structures of the different features represented. Therefore, we recover samples from each different set of features so as to faithfully represent the distribution of each individual set. 

In this process, we extract samples from each set $F_1, F_2, \ldots, F_p$ separately using a cluster-based strategy. We employ the k-means algorithm to create $\sqrt{n}$ clusters, getting the medoid of each cluster as a sample, where $n$ is the number of instances in the raw dataset $D$. We set the number of cluster to $\sqrt{n}$ since this is considered a good heuristic for the upper-bound number of clusters in a data set~\cite{413225}. After extracting the sample sets $S_1, S_2, \ldots, S_p$, we merge their indexes defining a unified set of indexes. Then we recreate the sets $S_1, S_2, \ldots, S_p$ to have the instances with the indexes contained in the unified set of indexes. Therefore, all sample sets have the same instances, which is mandatory for the sample visualization given that we visualize the combination of all features $\overline{R}$. Also we guarantee that the structures defined by the different types of features are represented by the samples. Notice that, the combined sample features $\overline{R}$ will have at most $\sqrt{n} \times p$ instances, enhancing the probability of having samples that represent the distribution of each individual set of features while not hampering the computational complexity of the overall process since $p\ll\sqrt{n}$.

After recovering the samples, we map them to a common $m$-dimensional space, obtaining their vectorial representation $R_1,R_2,\ldots,R_p \in \mathbb{R}^m$ so that we can combine them to obtain $\overline{R} \in \mathbb{R}^m$ (for the sample visualization). In this process, each set of samples $F_i$ is mapped to $\mathbb{R}^m$ preserving as much as possible the distance relationships in $F_i$. We do this by minimizing 

\begin{equation}
E_{st}(F_i) = \frac{1}{|F_i|^2}\sum_{i}^{|F_i|}\sum_{j}^{|F_i|}\left(\delta(f^i_i,f^i_j) - ||r^i_i-r^i_j||\right)^2 
\label{eq:stress}
\end{equation}
where $f^i_i$ and $f^i_j$ are instances in $F_i$, $\delta(f^i_i, f^i_j)$ is the distance between them, and $r^i_i$ and $r^i_j$ are the vectorial representations in the $m$-dimensional space of $f^i_i$ and $f^i_j$, respectively. 

Besides preserving distance relationships, our mapping process aim to align the vectorial representations so that $r^i_i$ is placed as close as possible to $r^j_i~\forall~j~\in [1,p]$ without affecting the distance preservation of the individual mappings. This is necessary since the unified sample representation is calculated as a convex combination of these representations, that is, $\overline{R}=\alpha_1R_1,\alpha_2R_2,\ldots,\alpha_pR_p$, with $\sum\alpha_i=1$, and misalignments could result in meaningless unified representations. We first calculate the normalized average distance matrix $\overline{\Delta} = \frac{1}{p}\sum_i^p \Delta_{F_i}$ combining the distance matrices of all sets of features, where $\Delta_{F_i}$ is the distance matrix calculated from $F_i$. Then we map $\overline{\Delta}$ to the $m$-dimensional space using the Equation~(\ref{eq:stress}). The idea is to use this average representation as a guide to align the vectorial representations $R_1,R_2,\ldots,R_p$ minimizing

\begin{equation}
E_{al}(F_i) = \frac{1}{|F_i|^2}\sum_{i}^{|F_i|}\sum_{j}^{|F_i|}\left(d(\overline{r}_i,\overline{r}_j) - ||\overline{r}_i-r^i_j||\right)^2
\label{eq:alignment}
\end{equation}
where $d(\overline{r}_i,\overline{r}_j)$ is the distance between two instances of the average vectorial representation. 
 
Joining Equation~(\ref{eq:stress}) and (\ref{eq:alignment}) we define the function we optimize in our mapping process seeking to preserve, as much as possible, the distance relationships of the original features $F_1, F_2, \ldots, F_p$ in the vectorial representations $R_1,R_2,\ldots,R_p \in \mathbb{R}^m$ while aligning them. This function is given by

\begin{equation}
E(F_i) = \lambda\cdot E_{st}(F_i) + (1-\lambda)\cdot E_{al}(F_i)
\label{eq:mapping}
\end{equation}
where $\lambda$ is a used to control the importance of the distance preservation and the alignment to the produced vectorial representations. $\lambda$ is a a hyperparameter and can be changed to defined a good tradeoff between distance preservation and alignment.

To minimize Equation~(\ref{eq:mapping}) we use a stochastic gradient descent approach with a polynomial decay learning rate. We set the initial learning rate to $\gamma_0=0.1$ and the decay power to $\kappa=0.95$ following common choices found in the literature~\cite{Wilson:2001}. Algorithm~\ref{alg:dr} outlines our mapping process. Function \textsc{random}$(F_i,n)$ select $n$ samples from $F_i$ randomly and function \textsc{init}() initialize the mapping also randomly. We have tested a deterministic initialization using Fastmap~\cite{Faloutsos:1995:FFA:568271.223812} but the gain in quality does not justify the computational overhead. 
Notice that we normalize all features $f_i \in F_j,~\forall~i,j$ before this process, so that the Euclidean norm $||f_i||=1$. Given the triangular inequality property ($||f_i-f_j|| \leq ||f_i||+||f_j||$), this guarantees a upper limit for the maximum pairwise distance between features. Therefore the distances are in the same range despite the type of feature or its dimensionality, avoiding biasing the process towards the type of feature with the largest maximum distance. In addition, we define the desire dimensionality $m$  of the resulting mappings $R_1,R_2,\ldots,R_p\in\mathbb{R}^m$ as the largest intrinsic dimensionality of $F_1, F_2, \ldots, F_p$, calculated using the maximum likelihood estimation~\cite{Levina:2004}. Such dimensionality can also be defined by the user if the target dimensionality is known, such as, ${m=\{1,2,3\}}$ for visualization purposes.

\begin{algorithm}{}
\algrenewcommand\algorithmicindent{1.0em}%
\begin{algorithmic}
\State $\overline{\Delta} \leftarrow \frac{1}{k}\sum_i^k \Delta_{F_i}$ \Comment{calculate the average distance matrix}
\State $\overline{R} \leftarrow \textsc{mapping}(\overline{\Delta}, 1.0)$ \Comment{compute the dimensionality reduction of $\overline{\Delta}$}
\For{$F_i \in {F_1, F_2, \ldots, F_k}$}
  \State $R_i \leftarrow \textsc{mapping}(F_i, \overline{R}, \lambda)$
\EndFor
\vspace{0.5cm}
\Function{mapping}{$F$, $\overline{R}$, $\lambda$}
  \State $R \leftarrow \textsc{init()}$ \Comment{initialize the dimensionality reduction}  
  \For{$it=0~to~\Omega$}
    \State $\gamma \leftarrow \gamma_0 \times \left(1-\frac{it}{\Omega}\right)^\kappa $ \Comment{polynomial decay of the learning rate}   
    \State $F_{rand} \leftarrow \textsc{random}(F, \sqrt{|F|})$ \Comment{get $\sqrt{|F|}$ random samples from $F$}
    \For{$f_i \in F_{rand}$}
      \For{$f_j \in F$}
        \State $\nabla E_{st} \leftarrow \left(\delta(f_i,f_j) - ||r_i-r_j|| \right)\frac{(r_i-r_j)}{||r_i-r_j||}$
        \State $\nabla E_{al} \leftarrow 
        \left(d(\overline{r}_i,\overline{r}_j) - ||\overline{r}_i-r^j|| \right)\frac{(\overline{r}_i-r_j)}{||\overline{r}_i-r_j||}$
        \State $r_j \leftarrow r_j - \gamma\left(\lambda \cdot \nabla E_{st} + (1-\lambda)\cdot\nabla E_{al}\right)$
      \EndFor
    \EndFor
  \EndFor
  \State \Return $R$
\EndFunction
\end{algorithmic}
\caption{Algorithm for mapping different feature sets to a common vectorial space.}\label{alg:dr}
\end{algorithm}


\subsection{Weighted Feature Combination}

Given the samples vectorial representations $R_1,R_2,\ldots,R_p$ we build a set of functions using the process defined in~\cite{Joia:2011:LAM:2068462.2068681} to map each feature set $F_i$ into its vectorial representation ${V_i \in \mathbb{R}^m}$ preserving as much as possible the distance relationships while obeying the geometry define in $R_i$. In this process, each instance $f^i_j \in F_i$ is mapped to the $m$-dimensional space trough a orthogonal local affine transformation $T^i_j:\mathbb{R}^{q^i}\rightarrow\mathbb{R}^m$, where $q^i$ is the dimensionality of $F_i$.

The affine transformation $T^i_j(f)= fM+t$ associated to $f^i_j$ is defined so as to minimize:
\begin{equation}
\displaystyle\sum_k \beta_k\|T^i_j(s^i_k)-r^i_k\|^2
\label{eq:inicialenergy}
\end{equation}
where $\beta_k=\|s^i_k-f^i_k\|^{-2}$, with $s^i_k$ the original feature representation of the $k$-th sample in $S_i$.

Equation~(\ref{eq:inicialenergy}) can be re-written in the matrix form $\|D\left(AM-B\right)\|_F$, where $\|\cdot\|_F$ denotes the Frobenius norm, $D$ is a diagonal matrix with entries $D_{ii}=\sqrt{\beta_i}$, and $A$ and $B$ are matrices with the $j$-th row given by the vectors
$$
s^i_j - \frac{\sum_k \beta_k s^i_k}{\sum_k\beta_k}
\quad\text{and}\quad
r^i_j - \frac{\sum_k \beta_k r^i_k}{\sum_k\beta_k}
\text{, respectively.}
$$ Based on that, $M$ is computed as $M=UV$ where $U$ and $V$ are obtained from the singular value decomposition of $A^{\!\top}DDB=USV^{\!\top}$. Then the vectorial representation $v^i_j$ of $f^i_j$ is given by

\begin{equation}
\displaystyle v^i_j= \left(f^i_j - \frac{\sum_k \alpha_k s^i_k}{\sum_k\alpha_k}\right)M+\frac{\sum_k \alpha_k r^i_k}{\sum_k\alpha_k}
\end{equation}

Equation~(\ref{eq:inicialenergy}) is subject to $MM^{\!\top}=I$, which avoids scale and shearing effects, therefore preserving the distance relationships of the input features. Also, notice that the sample vectorial representations $R_1,R_2,
\ldots,R_p$ dictates the geometry of the embeddings $V_1,V_2,\ldots,V_p$. Since they are aligned by the mapping process defined in the previous section, the linear combination $V=\alpha_1V1, \alpha_2V_2, \ldots, \alpha_pV_p$ can be performed to obtain the final embedding $V$ that incorporates the structures defined by each set of features, weighted according to the user's point-of-view. For more information about this affine transformation and how the sample vectorial representation controls the final results, please refer to~\cite{Joia:2011:LAM:2068462.2068681}.

\begin{figure}[]
	\centering
	\includegraphics[width=.35\linewidth]{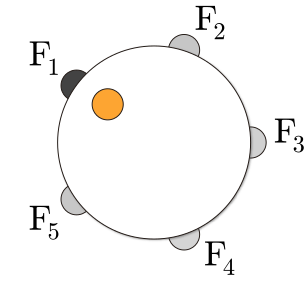}
	\caption{Feature Combination Widget. Using the orange ``dial'' users can control the contributions of the different types of features to the final feature combination.}
	\label{fig:widget}
\end{figure}

\subsection{Feature Combination Widget}

To visually support the feature sample combination, we create a widget inspired by the strategy presented in~\cite{10.1016/j.neucom.2014.07.072}. The idea is to position anchors (circles) representing each different set of features over a circumference, computing the weights $\alpha_1, \alpha_2, \ldots, \alpha_p$ according to their distances to a ``dial'' contained in the circumference. If $\tilde{f_i}$ are the coordinates of the anchor representing the feature $F_i$ on the plane and $\tilde{d}$ the coordinates of the ``dial'', the weight $\alpha_i$ related to $F_i$ is calculated as

\begin{equation}
\alpha_i = \left. 1 \middle/ \left(\sum_j^p\frac{(1+\|\tilde{f_i}-\tilde{d}\|)^2}{(1+\|\tilde{f_j}-\tilde{d}\|)^2}\right)\right.
\end{equation}

To help the perception of the weights, we change the transparency level of the anchors and fonts according to $\alpha_1, \alpha_2, \ldots,\alpha_p$. Figure~\ref{fig:widget} shows the combination widget. In this example, the ``dial'', in orange, is closer to the anchor representing the feature $F_1$, so the corresponding anchor is more opaque than the other anchors.

\section{Results and Evaluation}
\label{sec:res}

In this section, we evaluate our mapping and feature combination processes using different datasets aiming at showing that the sample manipulation effectively controls the complete feature fusion. Next, we describe the employed datasets, detail how we extract features, and present our quantitative and qualitative evaluation.

\subsection{Datasets} \label{sec:datasets}

We use five datasets in our tests, named \textbf{STL-10}~\cite{Coates:2011}, \textbf{Animals}~\cite{Lampert:2009}, \textbf{Zappos}~\cite{Yu:2014}, \textbf{CIFAR-10}~\cite{Krizhevsky:2009} and \textbf{Photographers}~\cite{Thomas:2016}. These datasets come from a variety of different domains. The \textbf{STL-10} consists of $13,000$ images split into $10$ classes of different objects. Similarly, \textbf{CIFAR-10} contains $60,000$ images of $10$ commonly seen object categories (e.g., animals, vehicles, and such) in lower resolution. The \textbf{Animals} dataset is more specific and it is composed of $30,475$ images of animals in $50$ categories. \textbf{Zappos} is a dataset for shoes with $50,025$ images from \url{Zappos.com} split into $4$ shoe categories.  Finally, the \textbf{Photographers} consists of $181,948$ photos taken by $41$ well-known photographers. Table~\ref{tab:datasets} summarizes the datasets, showing the number of instances and classes.


\begin{table}[!h]
    \centering
    \caption{Datasets employed in the evaluations. We report its size, number of classes, and intrinsic dimension. }
    \label{tab:datasets}
    \footnotesize{\begin{tabular}{|c|c|c|}
            \hline
            \textbf{Name}  & \textbf{Size} & \textbf{Classes} \\ \hline \hline
            STL-10                    & 13,000 & 10\\ 
            Animals                   & 30,475 & 50  \\ 
            Zappos                    & 50,025 & 4 \\ 
            CIFAR-10                  & 60,000 & 10 \\ 
            Photographer              & 181,948 & 41 \\ 
            \hline
        \end{tabular}}
\end{table}

\subsection{Features}\label{sec:features}

We use $4$ distinct methods to extract features, representing low-level and high-level image components. Low-level means that the dimensions of the feature vector has no inherent meaning, but represent a basic understanding of the image such as edges or color. High-level features have semantic meaning. For example, they denote the presence of an object or not in the image.

For the low-level features, we represent (1) color with LAB color histogram; (2) texture with Gabor filters~\cite{Chen:2004} with $8$ orientations and $4$ scales; and (3) shape with HoG technique~\cite{Dalalhog} with a window size of $8$. For the high-level, we extract deep-features from the \textit{pool5} layer using a pre-trained CNN CaffeNet~\cite{Jia2014}. This network was trained on approximately $1.3M$ images to classify images into $1,000$ object categories. 

We believe that these features are discriminative for our datasets. For example, we can differentiate a leopard from a panda using a texture extractor. Texture can identify spots in leopard, and differentiate them from other animals. Similarly, color features can be helpful to recognize pandas, where the more common colors are black and white. Also, HOG is helpful to differentiate the type of animals by their shape, e.g., quadrupeds from birds. Finally, object recognition can complement the HOG descriptor. These examples can be generalized to other datasets as well. 

\subsection{Quantitative Evaluation}\label{sec:qtevaluation}
 
 \begin{figure*}[]
 	\centering
 	\includegraphics[width=\linewidth]{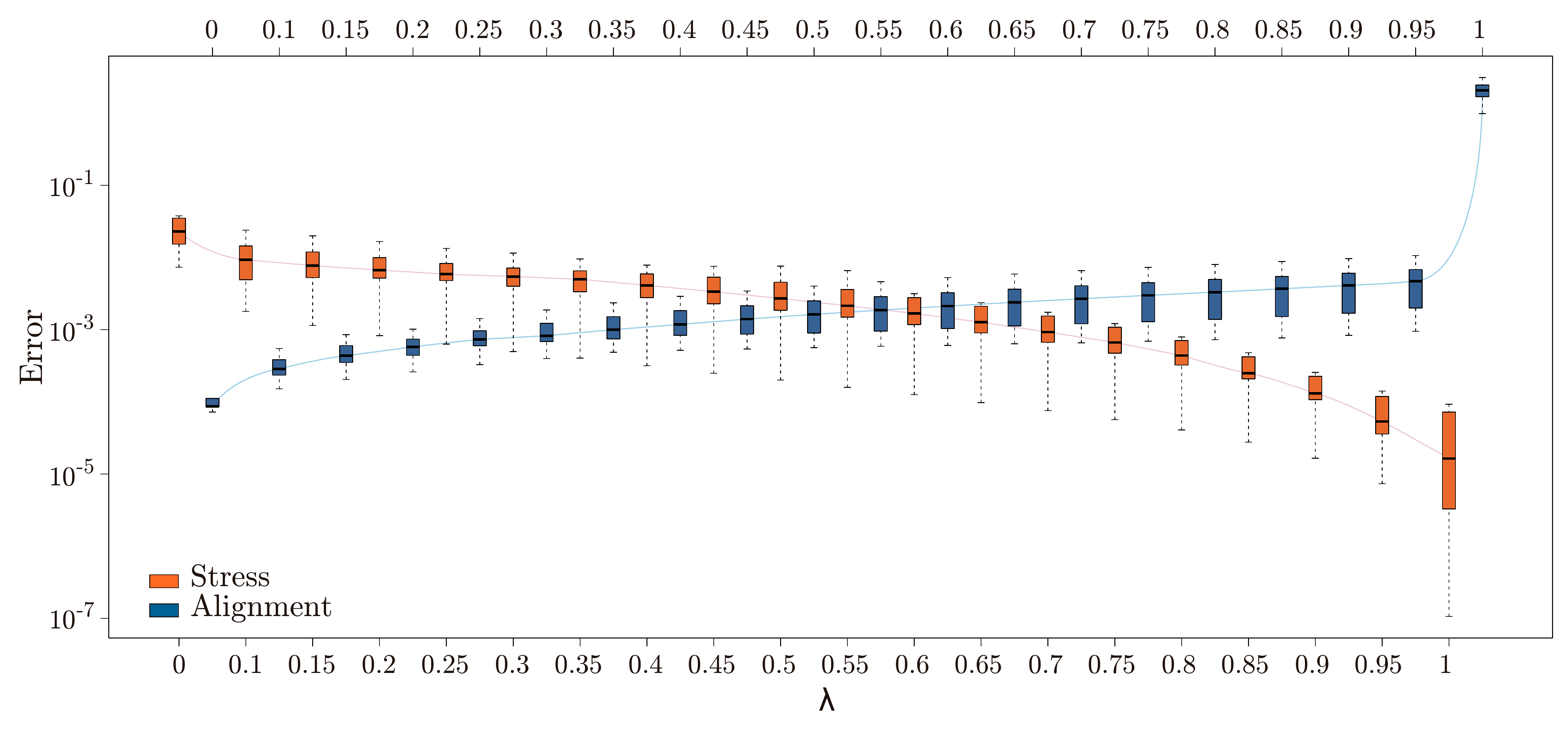}
 	\caption{Comparison for distance preservation and alignment error varying $\lambda$. The best trade-off is achieved in the range $[0.45, 0.65]$. The lines connect the average values of the boxplots. }\label{fig:lambda-eval}
 \end{figure*}

To confirm the quality of our approach, we quantitatively evaluate our mapping and feature combination processes. For the mapping process evaluation, the five datasets of Table~\ref{tab:datasets} are sampled $10$ times randomly reducing them to $5\%$ of their original sizes. We sample the data since we cannot execute the mapping process with large datasets since its memory footprint is O($n^2$). Due to the random initialization (see Algorithm~\ref{alg:dr}), we repeat the mapping process test $15$ times. Each different feature from the dataset has its own dimensionality. To ensure a common dimensional space, we calculate the intrinsic dimensionality for each of them and choose the smallest value. This value is used to do the mapping. The minimum values of intrinsic dimensionality are $57$, $71$, $91$, $41$, and $83$ for STL-10, Animals, Zappos, CIFAR-10, and Photographer datasets, respectively.

We use stress and alignment error to evaluate the mapping process (see Equation~(\ref{eq:stress}) and Equation~(\ref{eq:alignment}), respectively). We summarize our results in Figure \ref{fig:lambda-eval} varying the value of $\lambda$. The stress boxplots (in orange) decrease as $\lambda$ increases. On the other hand, alignment boxplots (in blue) have the opposite behavior. This is the expected outcome since larger values of $\lambda$ preserve the distance relationships, whereas small values align the data.

Setting $\lambda = 1$ preserves as much as possible the original distance relationships. This is reflected on a average stress of $E_{st}= 0.0009$, but it does not ensure a good alignment (average alignment of $E_{al} = 2.0343$). 
On the other hand, $\lambda = 0$ delivered almost a perfect alignment (average alignment of $E_{al} = 0.0001$), but it does not enforce the distance preservation (average stress of $E_{st}= 0.0345$). In this paper, we are interested in the best trade-off between distance preservation and alignment so that the alignment is obtained without penalizing the overall distance preservation of the mappings. According to our experiments, we achieved this in the range $\lambda = [0.45, 0.65]$, where both stress and alignment errors are nearly $0$ for our experiments (see Figure \ref{fig:lambda-eval}).

\begin{figure*}[]
    \centering
    \includegraphics[width=0.85\linewidth]{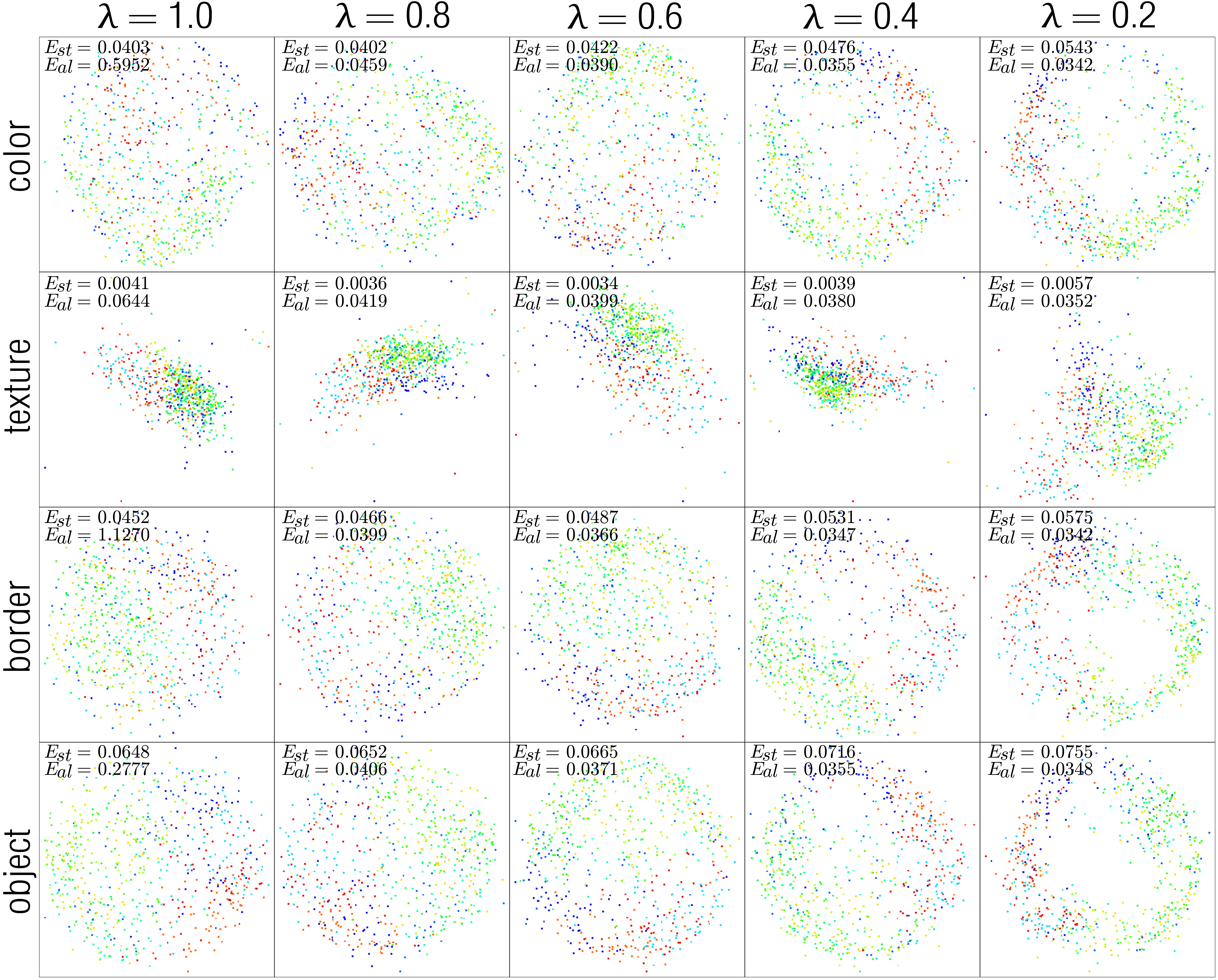}
    \caption{Resulting 2D mapping process for the \textbf{STL-10} dataset. As $\lambda$ decreases, the features get more aligned (See column 5). Top-left numbers correspond to stress and alignment error.}\label{fig:map-stl}
\end{figure*}

\begin{figure*}[]
    \centering
    \includegraphics[width=0.85\linewidth]{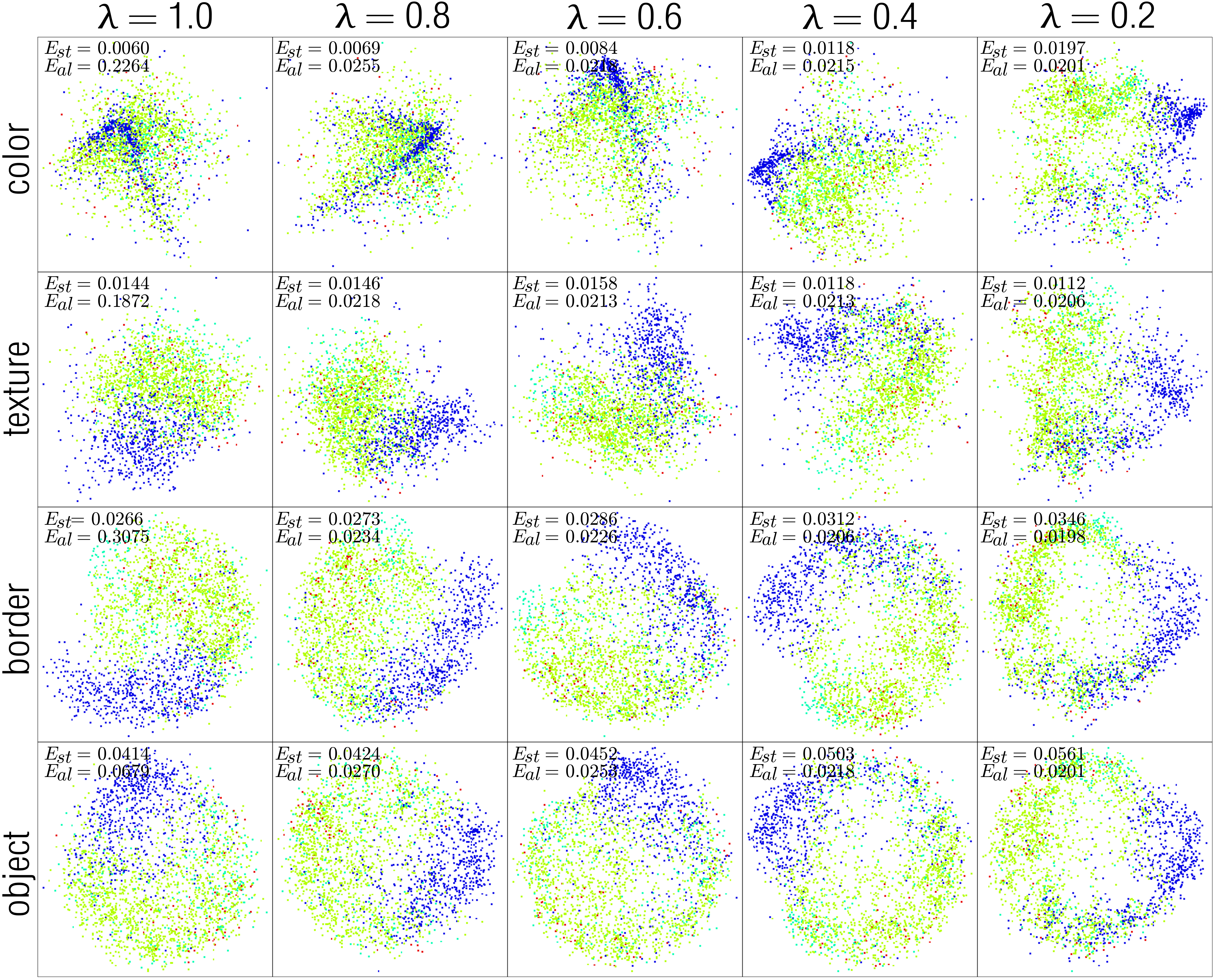}
    \caption{Resulting 2D mapping process for the \textbf{Zappos} dataset. As $\lambda$ decreases, the features get more aligned (See column 5). Top-left numbers correspond to stress and alignment error.}\label{fig:map-zappos}
\end{figure*}

\begin{figure*}[]
    \centering
    \includegraphics[width=0.85\linewidth]{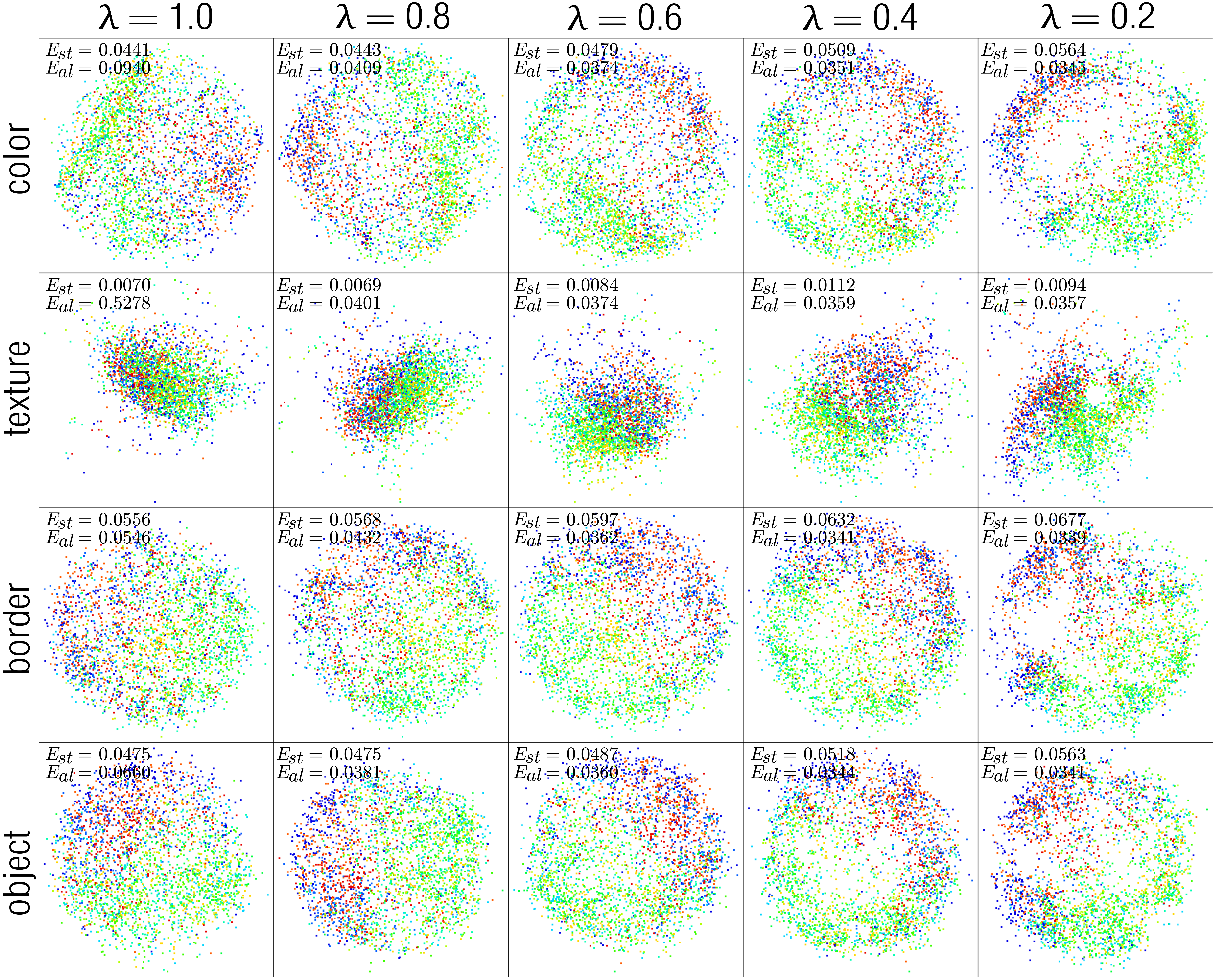}
    \caption{Resulting 2D mapping process for the \textbf{CIFAR} dataset. As $\lambda$ decreases, the features get more aligned (See column 5). Top-left numbers correspond to stress and alignment error.}\label{fig:map-cifar}
\end{figure*}

For a qualitative evaluation, we generate two-dimensional representations of the samples using our mapping process setting the target dimensionality to two. We show the results for the \textbf{STL-10}, \textbf{Zappos} and \textbf{CIFAR-10} datasets in Figures~\ref{fig:map-stl}, \ref{fig:map-zappos}, and \ref{fig:map-cifar}, respectively. In these figures, the points are colored according to image classes. The stress and alignment error values are shown on the top-left corner of each scatterplot. To show the influence of different $\lambda$ in the mapping process, we vary it in the range $[1.0,0.2]$. The first column shows the result produced using $\lambda=1.0$, best preserving the original distance relationship. Notice that the visual representations of each different feature are misaligned among themselves. The second column depicts results with $\lambda=0.8$. Now, the 2D mappings start to align (points of representing images of the same class are placed in close positions). We observe a small increase of the stress error, but the alignment error decreases considerably compared to the first column (see the second measure on the top-left corner). The same behavior is verified in the remaining columns. The last column aligns almost completely all features. As expected, as lambda decreases, the distance preservation also decreases (stress increases), and the alignment improves (alignment decreases). However, the stress changes are minimal.  Hence, our approach is capable of making a good alignment between features whereas preserving distance relationships. Similar behavior can be observed in Figure~\ref{fig:map-zappos} and Figure~\ref{fig:map-cifar}.

For the feature combination, we assess the degree the distance relationships of the sample are preserved into the feature fusion of the whole dataset, intending to demonstrate the effectiveness of the user sample manipulation to the produced dataset. In this evaluation, we first generate $30$ different weight combinations randomly summing up to $1$ and apply it to sample data. Then, we reuse these weights for the whole data fusion and measure if the distance relationships induced by the weights on the sample are presented in the whole dataset. We use the Nearest Neighbor Measure (NNM)~\cite{Cui:2006} in this analysis. 


NNM quantifies the similarity of each instance in the whole data with its nearest neighbor in the sampled data. NNM is given by Equation~\ref{eq:nnm}, where $D_i$ is the smallest distance among the $i-th$ instance in the complete dataset and the instances in the sample, and $N$ denotes the number of instances. The authors normalized each dimension of the data to the range $[0, 1]$. However, this results in the loss of the magnitude of the dimensions. So, we change the normalization per dimension by a unit vector normalization per instance to avoid such an effect. The output of NNM is in the interval $[0, 1]$ with larger values indicating better results.

\begin{equation}
    NNM = 1.0 - \frac{\sum^N_i D_i}{N}
    \label{eq:nnm}
\end{equation}

\begin{figure}[]
    \centering
    \includegraphics[width=0.95\linewidth]{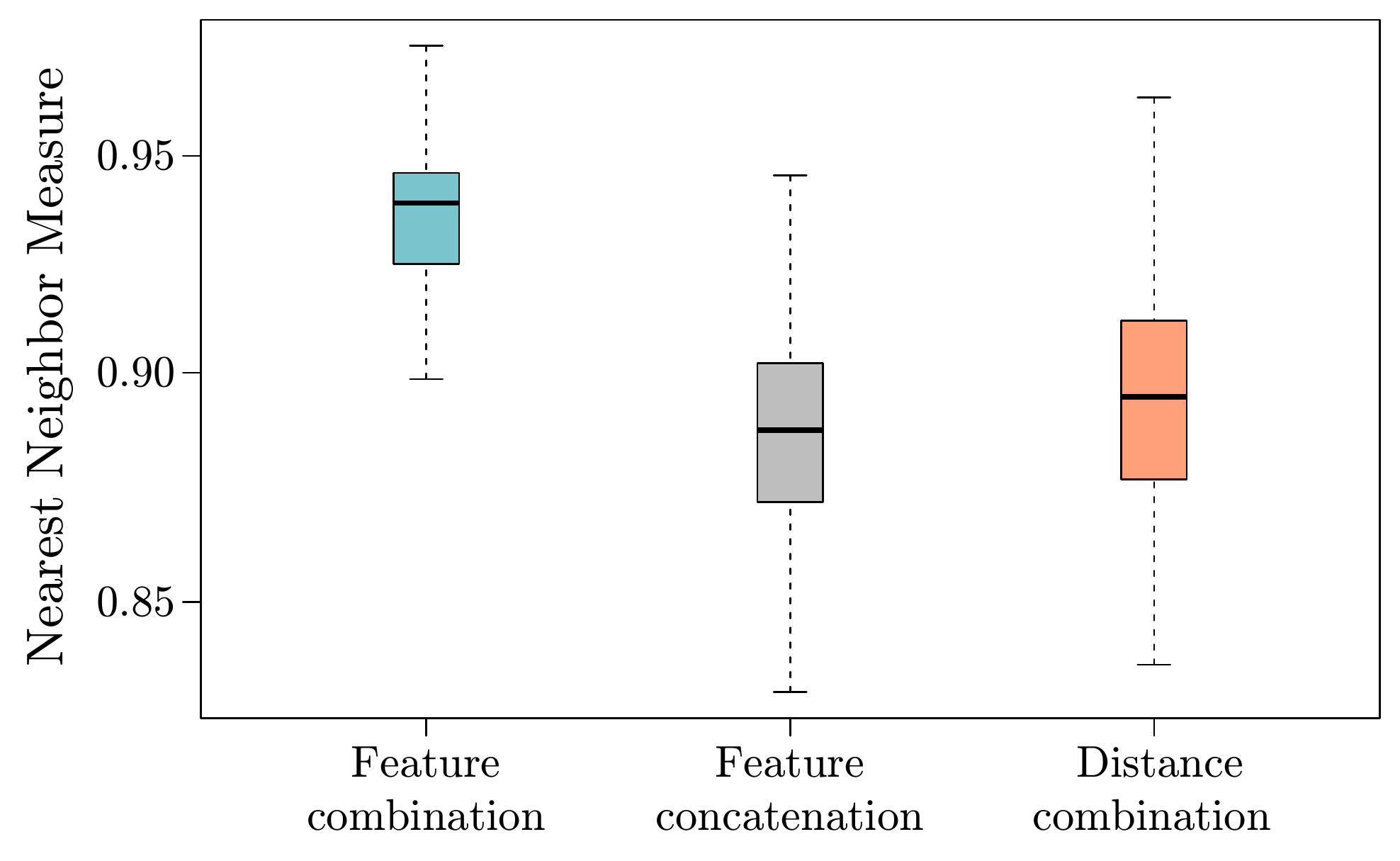}
    \caption{NNM evaluation. We compare our approach of user-guided feature fusion (light green box), with two baselines: feature concatenation, and feature distance combination.  Our feature fusion strategy surpasses current state-of-the-art strategies, indicating that the similarity patterns observed in the sample data combination are preserved on the complete dataset combination.
    }\label{fig:nnm-eval}
\end{figure}

We compare the NNM values of our feature fusion with two baselines: feature concatenation and distance fusion (see Section~\ref{sec:related}). Boxplots in Figure~\ref{fig:nnm-eval} show that our approach outperforms the other two baselines by at least $5\%$. The mean value for our method is $0.9365$, and the baselines achieve $0.8877$ and $0.8958$, respectively. Hence, our method preserves more accurately the data distribution of the sample in the whole dataset fusion.

\subsection{Qualitative Evaluation}\label{sec:qlevaluation}

Besides de quantitative evaluation, we also present an example based on projections for qualitative evaluation. The reasoning is to project the complete combined dataset ($E$), showing that the patterns observed in the sample projection ($R$) are preserved on the complete projection. In this example, we use our approach to explore large photo collections considering different user perspectives about similarity among images. We use the \textbf{photographers} dataset. In addition to the features described on Section~\ref{sec:datasets}, we create a new set of features to describe each photographer. We use Wikipedia articles about each photographer and construct a bag-of-words vector to represent them. Photos of the same photographer share the same feature vector, and the similarity among photos is defined as the similarity between texts describing the photographers.

\begin{figure*}[h]
\centering
\subfigure[]{\includegraphics[width=.45\linewidth]{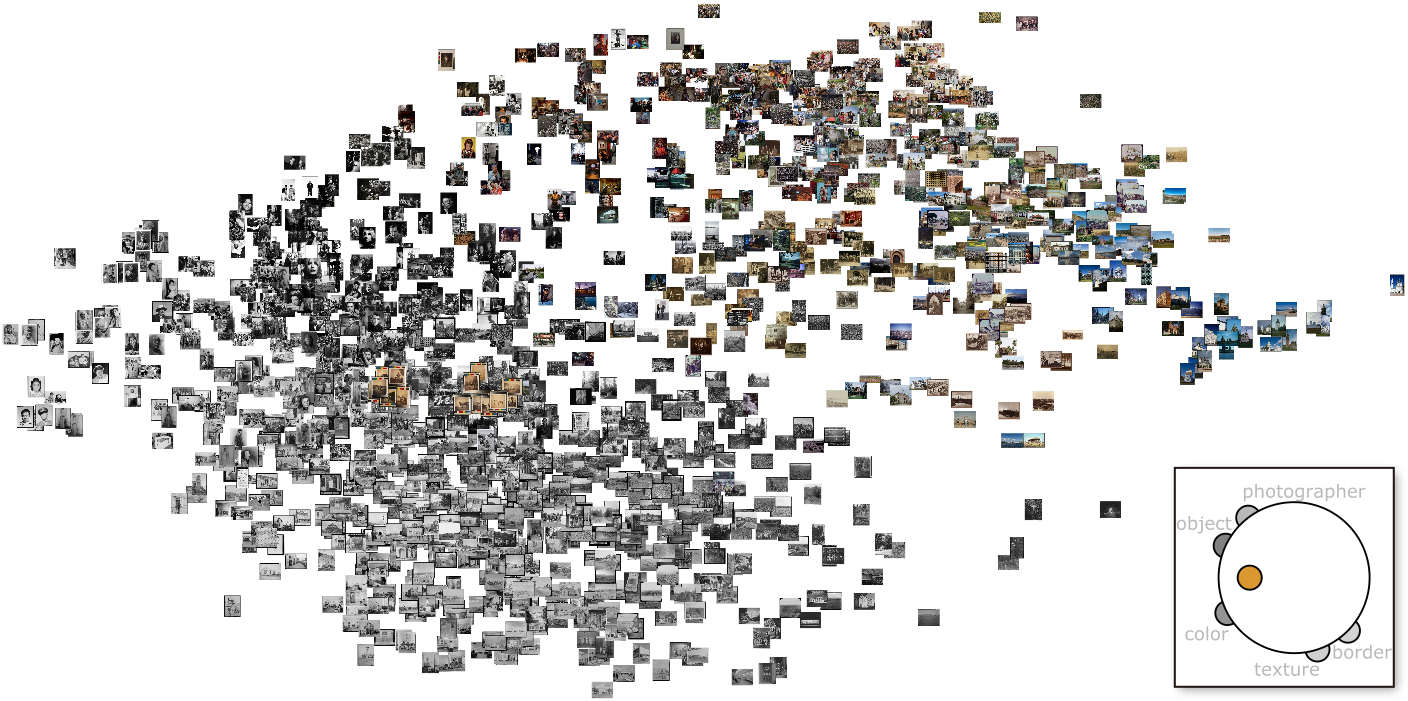}\label{fig:con1}}\quad
\subfigure[]{\includegraphics[width=.45\linewidth]{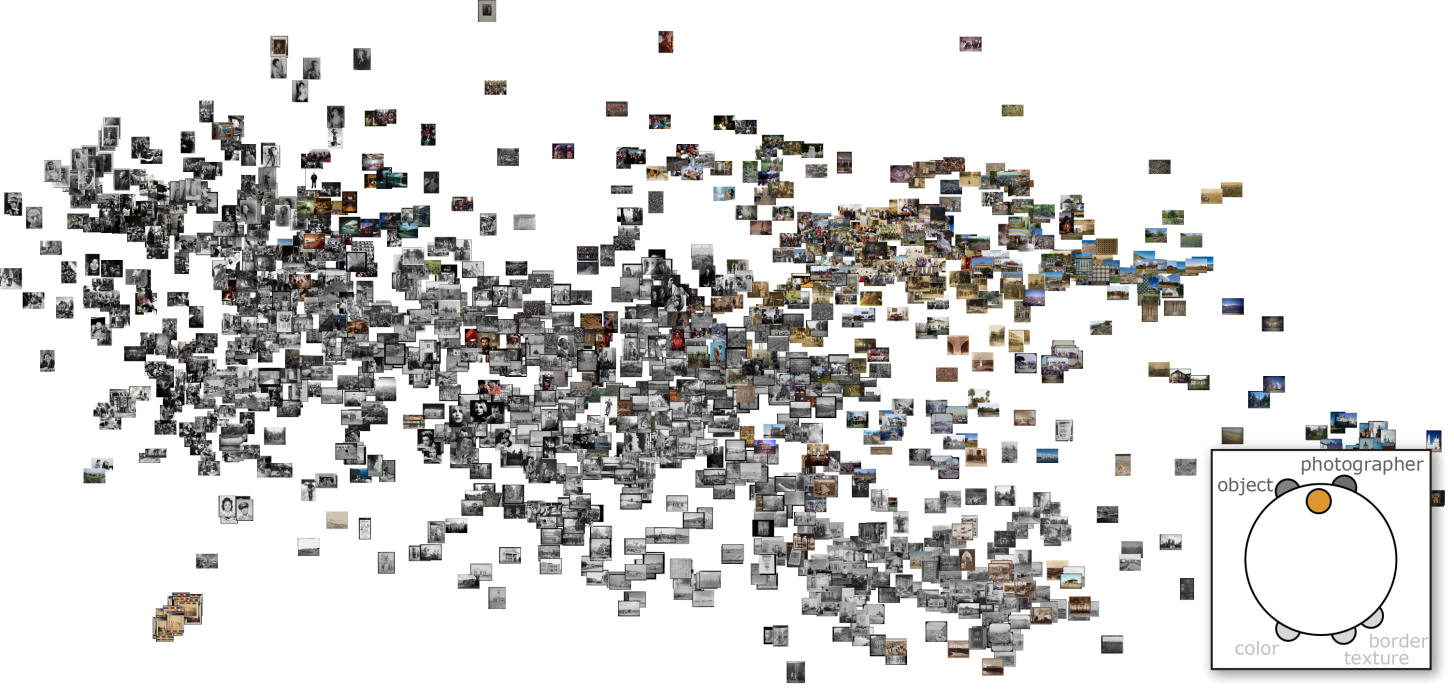}\label{fig:con2}}
\subfigure[]{\includegraphics[width=.45\linewidth]{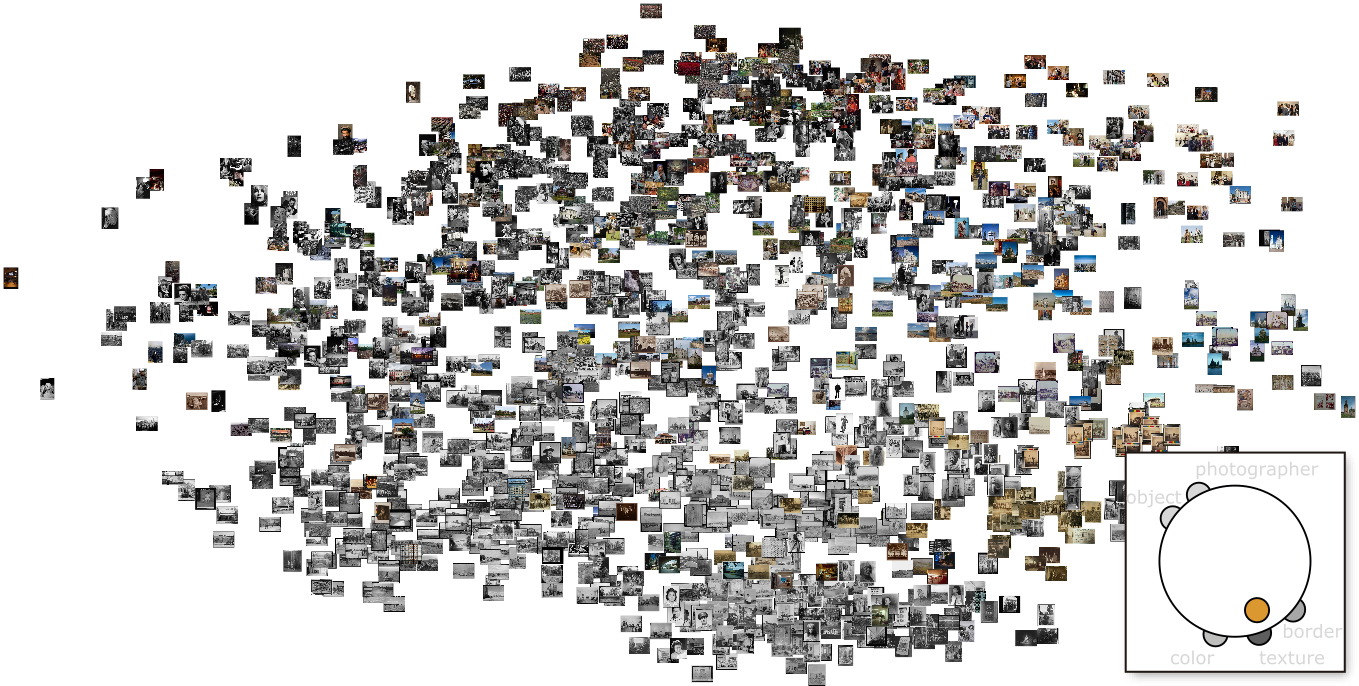}\label{fig:con3}}
\caption{User-defined similarity configurations. Based on a small sample, users can interactively combine different features seeking for the combination that best approaches a particular point of view. This combination is then propagated to the entire dataset for a complete projection.}
\label{fig:configurations}
\end{figure*}

As explained before, based on a sample, using our approach users can combine different features employing the combination widget (see Figure \ref{fig:widget}) until the sample visualization reflects a particular understanding regarding the similarity among photos. Figure~\ref{fig:configurations} shows three different combinations. The first (Figure~\ref{fig:configurations}(a)) provides more importance to color and objects contained in photos and little importance to information about photographers. The second (Figure~\ref{fig:configurations}(b)) is defined taking the idea of photographic style from~\cite{Thomas:2016}, fusing objects and Wikipedia features. Finally, the third (Figure~\ref{fig:configurations}(c)) shows the result of combining texture, borders and a little amount of color. 

\begin{figure*}[]
\centering
\includegraphics[width=\linewidth]{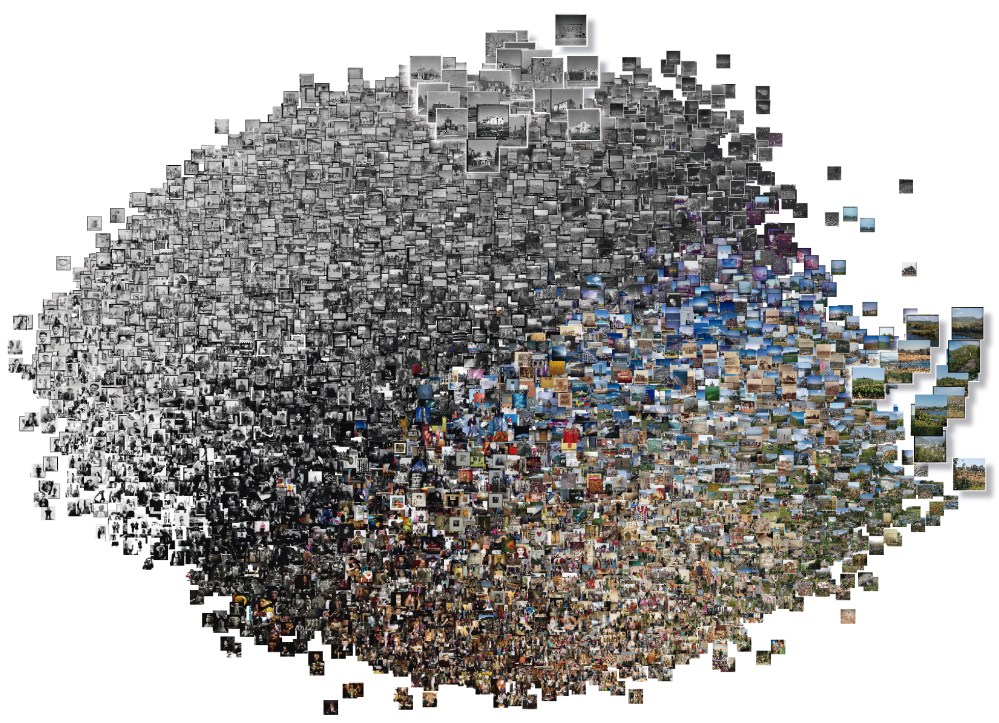}
\caption{Photographers dataset projection using the weight combination of Figure~\ref{fig:configurations} (a). Since a larger weight is assigned to the color feature, a clear global separation between black-and-white and colorful photos can be observed. This configuration also considers presence of objects and photographer information.}\label{fig:conf1}
\end{figure*}

Once the feature combination has been defined reflecting the users' point of view, a projection representing the complete photo collection is constructed. Figure~\ref{fig:conf1} shows the produced layout using the weights defined on Figure~\ref{fig:configurations}(a). In this figure, since the color is an important feature, we observe a clear separation between back-and-white and colorful images. Also, given the weight assigned to the features representing objects, it is possible to notice a separation among photos of people, landscapes, houses in certain regions of the figure.  We zoom in two small portion of the projection (at the top and at the right side) to show this effect. On the colored images (right), we observe images with sky and forest. On the gray images (top), we observe houses, sky, and forest.

\begin{figure*}[]
    \centering
    \includegraphics[width=\linewidth]{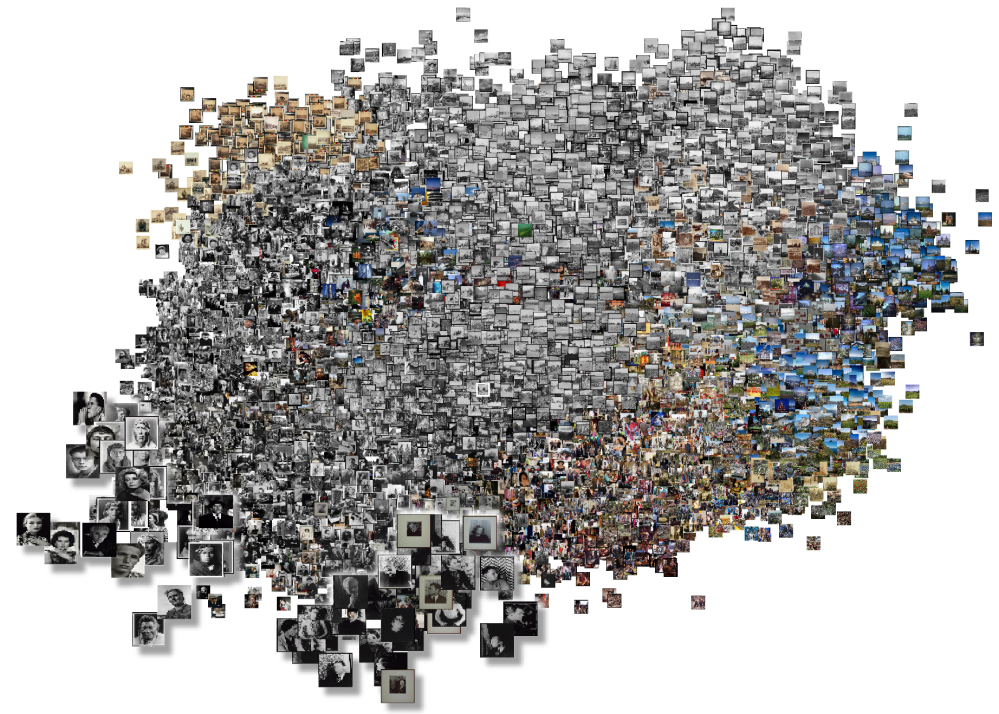}
    \caption{Photographers dataset projection using the weight combination of Figure~\ref{fig:configurations} (b). A larger weight is assigned to the object and photographers features. Photos with similar visual features are grouped. The zoom-in region (bottom-left) shows photos of well-known photographers that share similar styles (portraits photos).
    }
    \label{fig:conf2}
\end{figure*}

Figure~\ref{fig:conf2} depicts the final projection using the weights defined on Figure~\ref{fig:configurations}(b). In this figure, we zoom-in a region on the bottom-left. We mainly find portrait images in the zoomed region. Remember that in this weight combination, our goal was to represent the photographic style. The selected photos are from two well-known photographers, Van Vechten and Curtis, that mostly work with portraits, presenting similar styles~\cite{Thomas:2016}. These examples qualitatively attest that the similarity patterns observed on the sample projection are presented on the complete projection, corroborating the quantitative results measured using the NNM index.

\section{User-guided Clustering}
\label{sec:app}

One of the most appealing application scenarios for our approach is to assist non-supervised strategies, such as clustering techniques. Clustering techniques seek to split sets of data instances into groups so that instances belonging to the same group are more similar to each other than to those in other groups. Therefore, clustering is a subjective task that depends on the way similarity is computed, and the ability to explicitly control and understand similarity is the benefit our approach offers.

\begin{figure*}[]
    \centering
    \includegraphics[width=\linewidth]{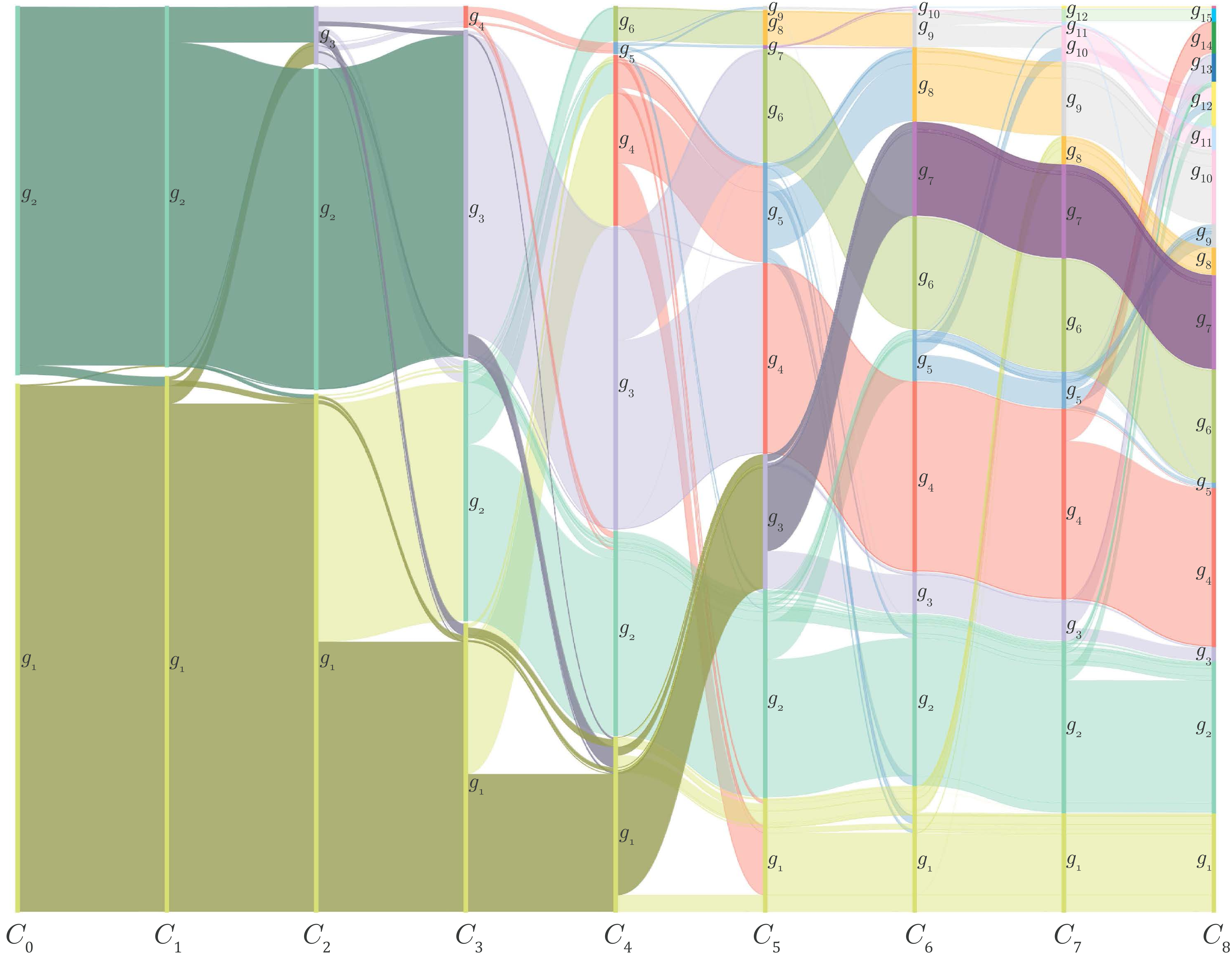}
    \caption{Using Parallel Sets to visualize cluster formation. The parallel sets visualization shows nine clusterings results computed using the Means-Shift algorithm. Axis $C_0$ and $C_8$ represent the clustering results for color and Wikipedia features, respectively. Intermediate clusterings denotes combinations of these features.} \label{fig:parallelset}
\end{figure*}

Following we present an example of using our approach to control clustering results of a sample of the \textbf{photographers}  dataset containing $7,800$ instances. In this example, we define different weights for features and observe how this influence the composed groups. In Figure~\ref{fig:parallelset} we analyze a transition between color and Wikipedia features. Color starts with weight $1$, and decreases to weight $0$ as Wikipedia weight increases from $0$ to $1$. 
We generate new fused features in each intermediate state. In each combination state, we compute clusters using the Mean-shift Algorithm~\cite{Comaniciu:2002}. We opt to use this algorithm because we do not need to provide the number of clusters as input, so the produced results directly reflect the provided similarity (or combination of features).

We display the different clustering configuration (for each combination) using the parallel sets~\cite{Kosara:2006}. On the parallel sets, the vertical axes represent different clusterings $C_k$, where $k$ indicates a different weight combination of features. All axes contain a set of groups where different colors represent different groups. Curves between axis $k$ and $k + 1$ are colored using the colors of the groups in $C_k$. This coloring scheme improves the perception of membership changes between different clusterings results. To reduce cluttering, we implement a simple filtering strategy to remove non-relevant curves. For each group in $C_k$, we evaluate how many instances from this group are redistributed in the groups in $C_{k + 1}$. If the quantity is less than a percentage threshold, the curves representing these instances are removed. This threshold is a user parameter and can be adjusted accordingly. 

Figure~\ref{fig:parallelset} shows parallel sets with $9$ axis representing clusterings results $(C_0, C_1,...,C_8)$ with a filtering threshold equals to $0.1$. $C_0$ axis represents the results for the color feature only (no Wikipedia feature is considered). It has two groups, one presenting colorful and the other black-and-white photos. $C_1$ shows fused features with $0.79$ weight to color and $0.21$ weight to Wikipedia. Most of the two groups presented in $C_0$ remain in $C_1$, but some instances change their membership.

\begin{figure*}[h]
\centering
\subfigure[$C_0$]{\includegraphics[width=.325\linewidth]{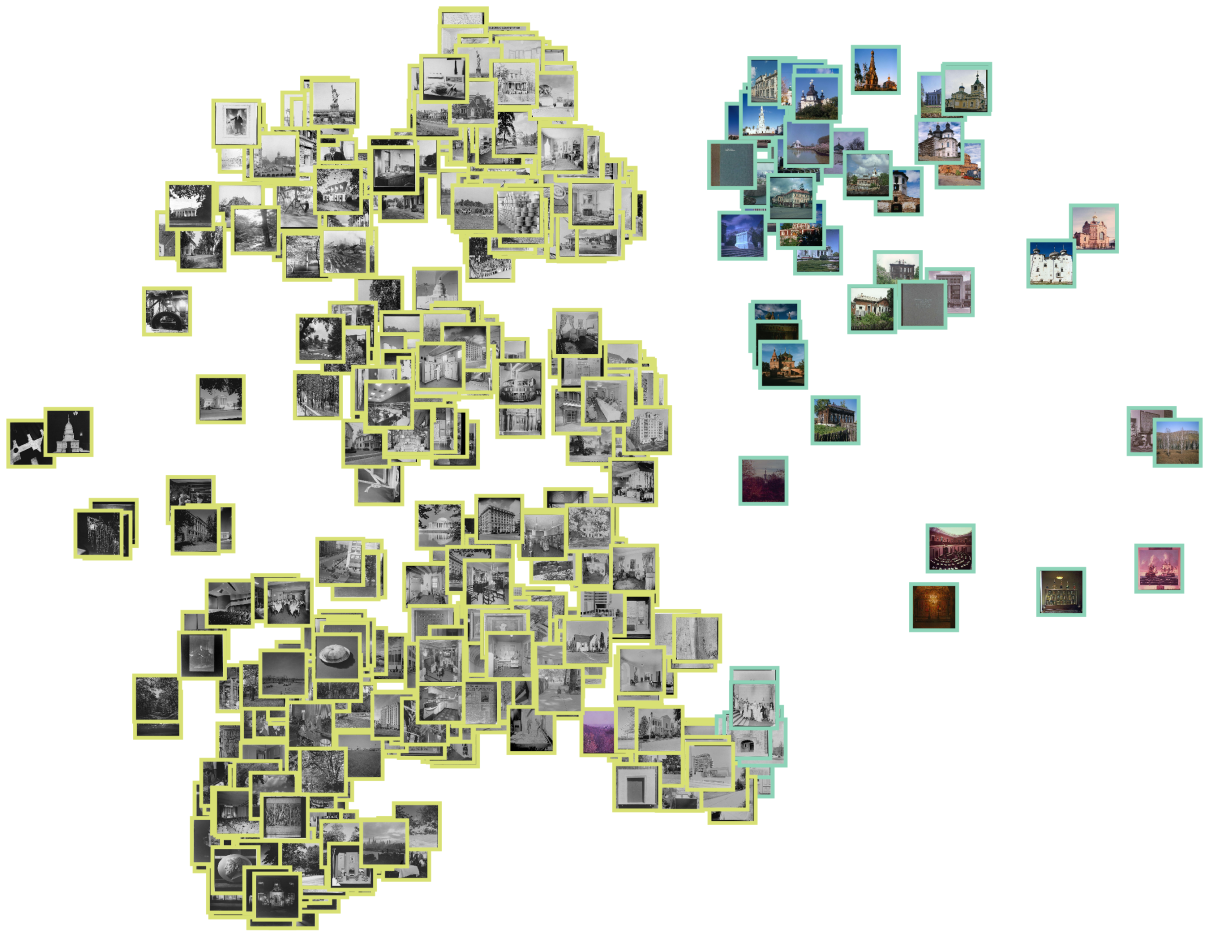}}
\hspace{1.3cm}
\subfigure[$C_3$]{\includegraphics[width=.325\linewidth]{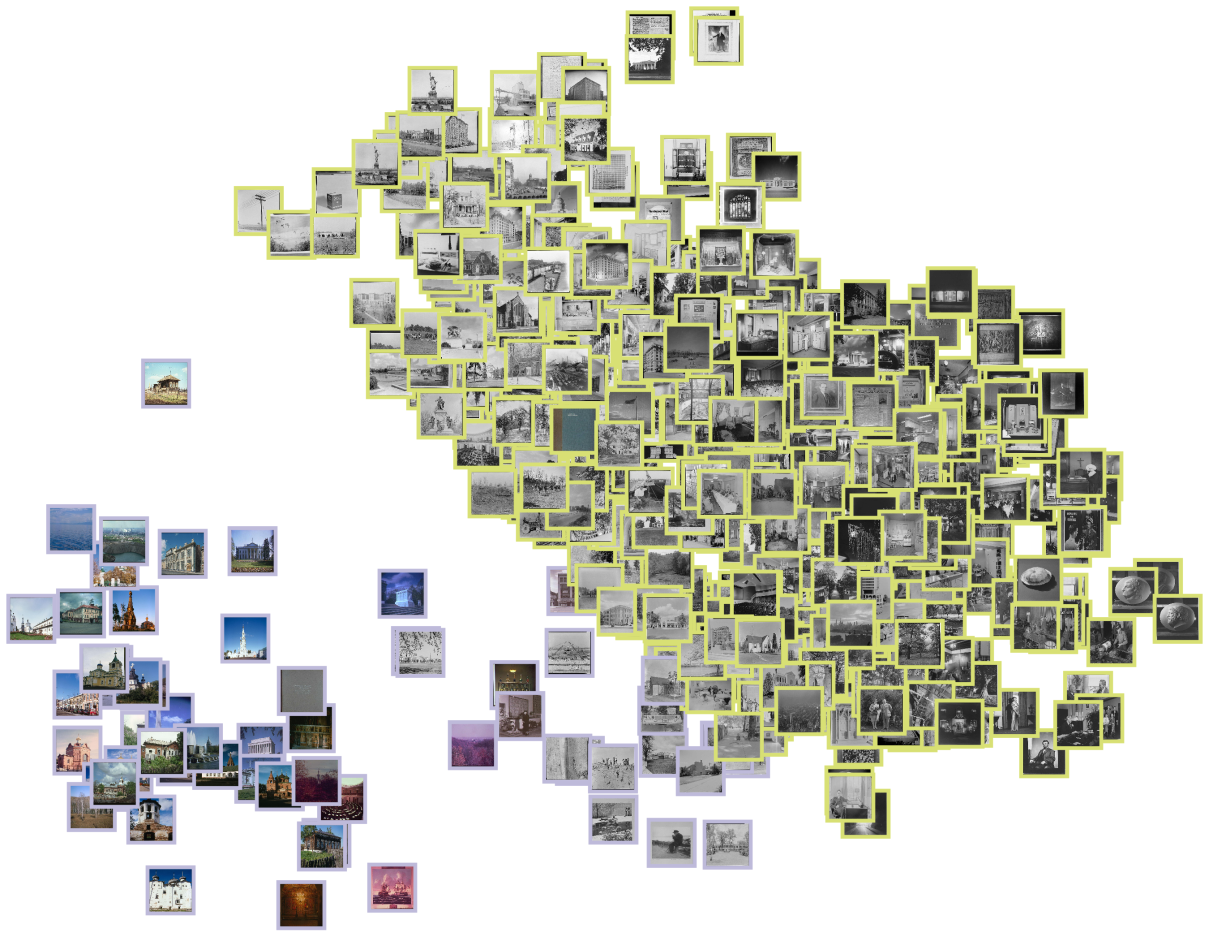}}\\
\subfigure[$C_4$]{\includegraphics[width=.325\linewidth]{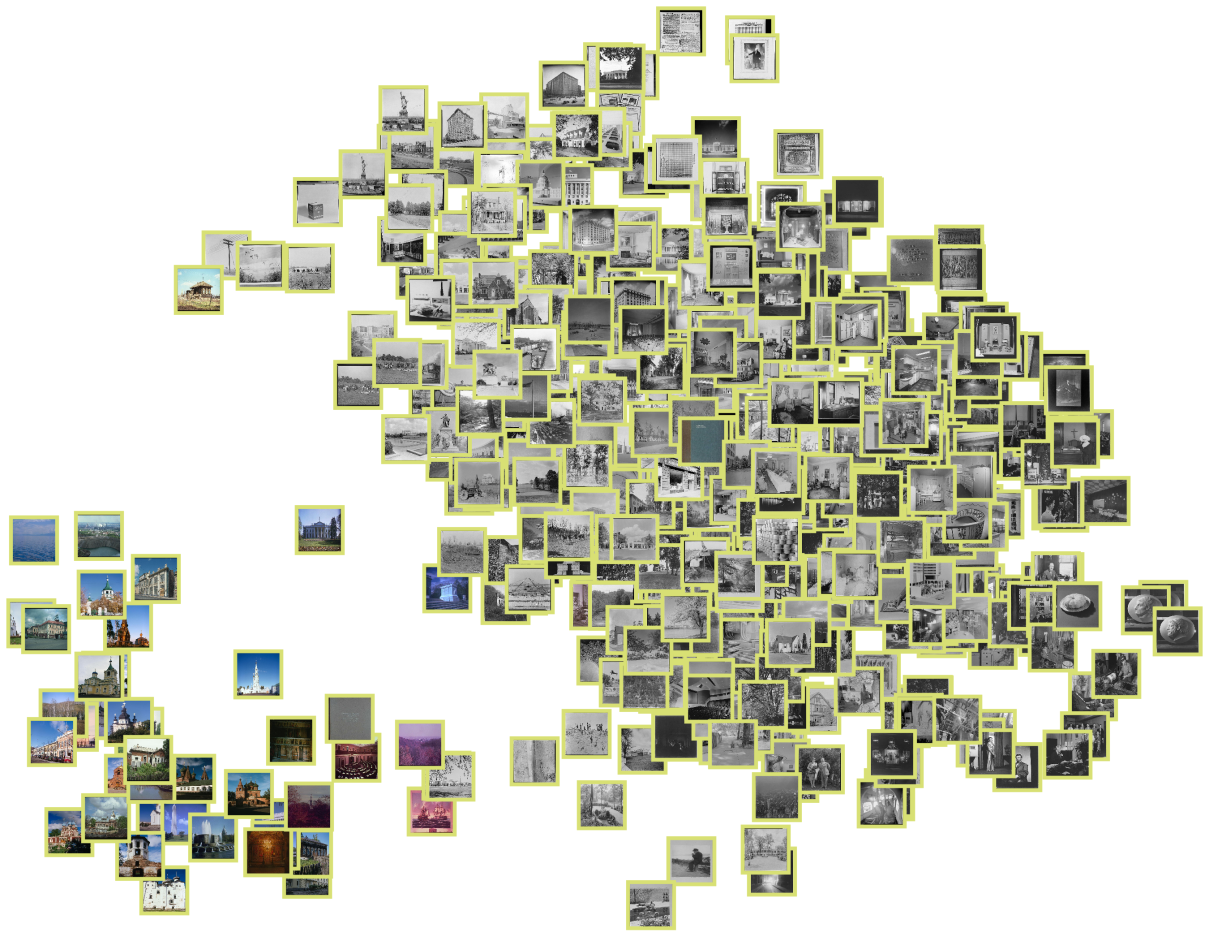}}
\hspace{1.3cm}
\subfigure[$C_8$]{\includegraphics[width=.325\linewidth]{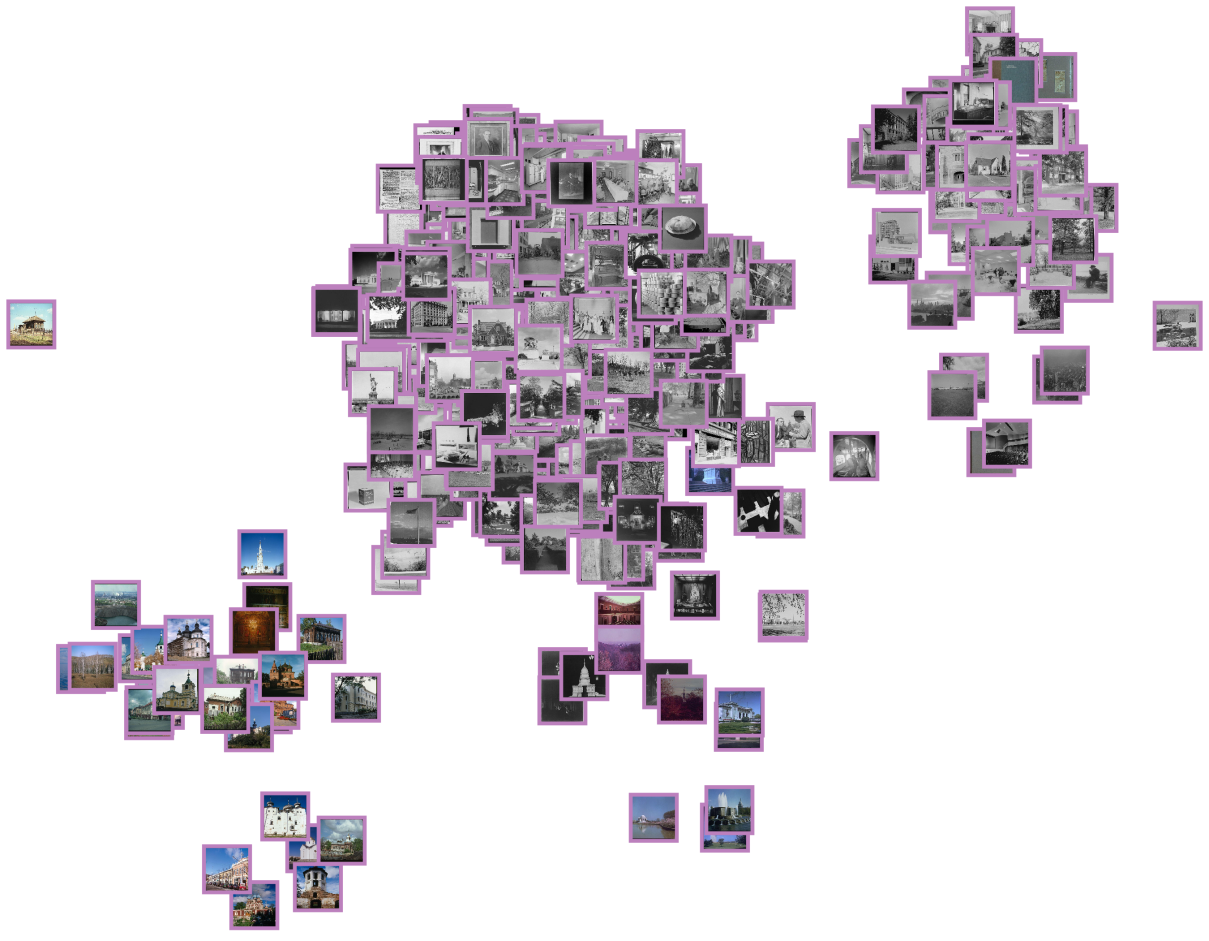}}
\caption{Projections produced by the Force-scheme technique for the purple group ($g_7$) of instances in $C_8$. The color of the border indicates the group the photo belongs to. The purple group instances are selected from $C_0$, $C_3$ and $C4$ and mapped in (a), (b) and (c), respectively. The projection of the purple group in $C_8$ is shown in (d). In (a), two groups are visible.  In (b), these groups are less separate. Both in (c) and (d), there are only one group according to the clustering technique. However, in (d) there is a small group inside this group that distinguishes three photographers with similar styles.}
\label{fig:projTransitions}
\end{figure*}

From $C_4$, more groups are composed, and the colorful and black-and-white photo division is lost. Finally, $C_8$ represents the clustering for Wikipedia feature only (no color features). Note that from clusterings $C_6$ to $C_8$, the groups are more stable, that is, most of the items in a certain group tend to be assigned to the same group as $k$ increases. In order to analyze the semantic meaning of the groups, we select the purple group ($g_7$) from $C_8$, and we check its correspondent instances backward. Photos of that group were taken by Brumfield, Gottscho, and Horydczak, which are three iconic American photographers. We map the data from $C_8,g_7$ to the visual space using the force-scheme technique~\cite{tejada:2003}. Figure ~\ref{fig:projTransitions}(d) shows the result where each photo border is colored with its group color. As can be observed, photos are similar in content and appearance. Brumfield, Gottscho, and Horydczak work is focused on architectural photography
We also observe that there is a mixture of colorful and black-and-white photos in this group. However, clustering $C_0$ shows a clear separation between these two types of photos (Figure~\ref{fig:projTransitions}(a)). Looking at the sequence of curves from clustering $C_8$ to $C_0$, it is possible to analyze the $g_7$ group, and when these photos are merged backward. We highlighted the path in the parallel set with darker colors for easy navigation.
From $C_8$ to $C_4$, the groups are stable. Instances of that groups are also projected and depicted in Figures~\ref{fig:projTransitions}(d) and \ref{fig:projTransitions}(c). In $C_4$, $g_1$ is formed by instances from $C_3$, $g_1$ and $g_3$ groups. Corresponding instances from $C_3$ are mapped in Figure~\ref{fig:projTransitions}(b). Note that in $C_3$, colorful and black-and-white photos are mixed.

\begin{figure*}[h]
	\centering
	\subfigure[$Tran_0$]{\includegraphics[width=.4\linewidth]{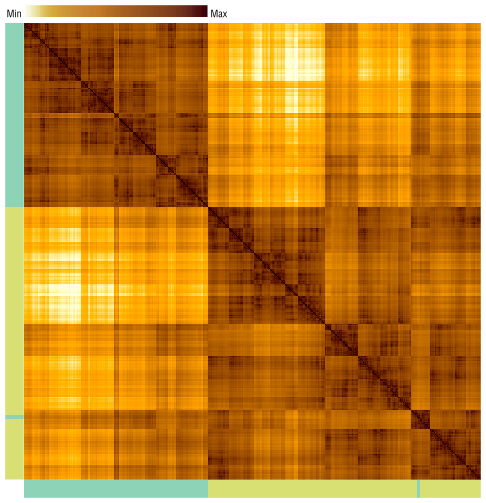}}
	\subfigure[$Tran_3$]{\includegraphics[width=.4\linewidth]{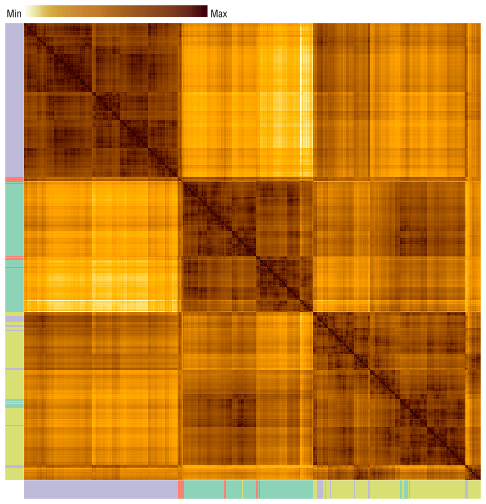}} \\
	\subfigure[$Tran_4$]{\includegraphics[width=.4\linewidth]{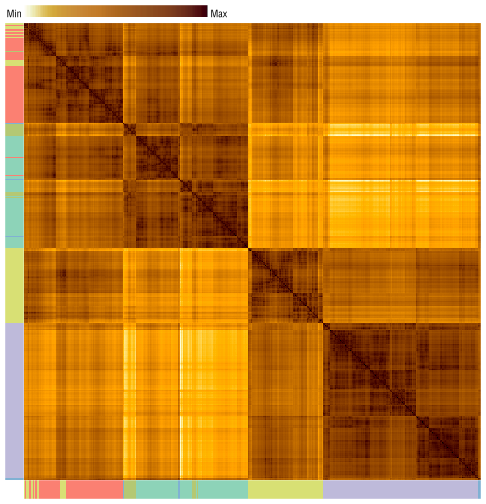}}
	\subfigure[$Tran_8$]{\includegraphics[width=.4\linewidth]{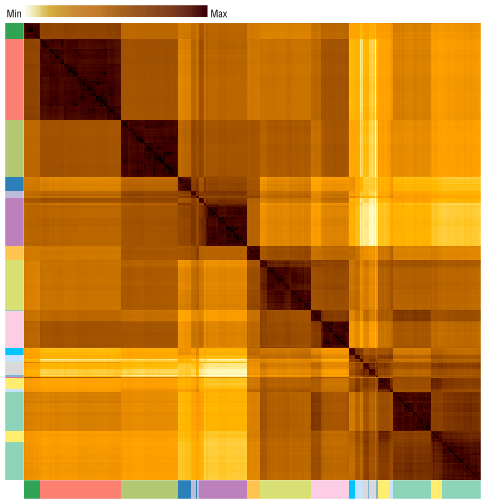}}
	
	\caption{Similarity Matrix for $4$ different weight combinations. $Tran_0$ represent color feature only, showing two groups (brown areas on the main diagonal). $Tran_8$ represent Wikipedia feature only and has several small groups in the diagonal and two major intersected groups. $Tran_3$ and $Tran_4$ have two major brown areas but with different sizes. These visualizations show how the different weight combinations influence the similarity calculation between instances, matching with the group formation presented by the parallel sets.}
	\label{fig:simMatrix}
\end{figure*}

Parallel sets are useful tools to show the difference between clustering results. However, they do not show the similarity relationships between instances. In order to explore clusters and the relationships between instances, we also visualize the pairwise dissimilarity matrix produce from a given feature combination as a heatmap. In our representation, similar items are rendered in brown colors, whereas dissimilar ones are rendered in pale orange colors. The order (rows and columns) of our representation is obtained by the position of the leaves in a dendrogram generated by the average linkage hierarchical clustering~\cite{Sokal:1958, Day:1984, Sander:2003}.

Figure~\ref{fig:simMatrix} shows dissimilarity matrices using the same weight combinations that generate $C_0$, $C_3$, $C_4$ and $C_8$ on parallel sets. In Figure~\ref{fig:simMatrix}(a) we can spot two groups (two brown areas on the main diagonal). The colored margins indicate the groups of the instances given by the clustering algorithm. Note there are some instances from the green group in the other group, denoting a potential problem with the clustering algorithm. Similar behavior can also be observed in Figure~\ref{fig:projTransitions}(a).  In Figure~\ref{fig:simMatrix}(b), the two major groups remain, but sub-groups can now be noticed inside the larger ones. Figure~\ref{fig:simMatrix}(c) also shows two big brown areas on the diagonal. However, these groups have the same size. In the previous matrix, one group is bigger than the other because the color feature has more weight and the dataset has more black-and-white than colorful photos. In Figure~\ref{fig:simMatrix}(c), Wikipedia feature begins to have more contribution in the combination process forming groups that groups photos according to style and color. Finally, in Figure~\ref{fig:simMatrix}(d), there are several groups on the main diagonal and two major groups that intersect. A possible explanation is that some photographers tend to shoot similar object categories, but they are from different schools of thought~\cite{Thomas:2016}. Looking at the purple part, we can see some sub-groups, each sub-group representing a photographer with a similar style. These sub-groups are also shown in Figure~\ref{fig:projTransitions}(d).


\section{Conclusion}
\label{sec:conclusion}

In this paper, we proposed a novel approach for feature fusion that successfully allows users to incorporate knowledge into the fusion process. It is a two-step strategy where, starting from a small sample of the input data, users can easily test different feature combinations and check in real-time the resulting similarity relationships. Once a combination that matches the user expectation is defined, it is propagated to the whole dataset through an affine transformation. Our experiments show that the complete dataset combination preserves the similarities from the sample configuration, providing our approach as a very flexible mechanism to assist the feature fusion process. 

We have applied the proposed feature fusion approach to allow users to control and understand the results of clustering techniques. Clustering is one of the most attractive application scenarios for our approach given the subjectiveness involved in unsupervised tasks. Currently, visualization assisted clustering techniques only allow to add user knowledge by changing techniques parameters~\cite{Kwon:2018, kern:2017, bruneau:2015, Wolfgang:2018}. Enabling users to guide the input feature configuration renders a much flexible control since users can explicitly steer the semantics of the input data and the similarity relationships (e.g.,  images are similar due to the color vs. images are similar due to the presence of objects), consequently controlling the reason for the cluster formation while allows an easy interpretation of the composed groups.



\bibliographystyle{SageV}
\bibliography{bibliography}
\end{document}